\documentclass{article}

\usepackage[english]{babel}

\usepackage[letterpaper,top=2cm,bottom=2cm,left=3cm,right=3cm,marginparwidth=1.75cm]{geometry}

\usepackage{amsmath}
\usepackage{graphicx}
\usepackage[colorlinks=true, allcolors=blue]{hyperref}
\usepackage{multirow}
\usepackage{amssymb}
\usepackage{authblk}  

\title{Precise Drive with VLM: First Prize Solution for \\ 
PRCV 2024 Drive LM challenge}

\author{
Bin Huang$^{*}$ , 
Siyu Wang$^{*}$, 
Yuanpeng Chen$^{*\dagger}$, 
Yidan Wu$^{*}$, \\
Hui Song, 
Zifan Ding, 
Jing Leng, 
Chengpeng Liang, 
Peng Xue, \\
Junliang Zhang, 
Tiankun Zhao
}
\affil{Hozon New Energy Automobile Co., Ltd}
\date{}

\begin{document}
\maketitle

\begin{abstract}
This technical report outlines the methodologies we applied for the PRCV Challenge, focusing on cognition and decision-making in driving scenarios. We employed InternVL-2.0, a pioneering open-source multi-modal model, and enhanced it by refining both the model input and training methodologies. For the input data, we strategically concatenated and formatted the multi-view images. It is worth mentioning that we utilized the coordinates of the original images without transformation. In terms of model training, we initially pre-trained the model on publicly available autonomous driving scenario datasets to bolster its alignment capabilities of the challenge tasks, followed by fine-tuning on the DriveLM-nuscenes Dataset. During the fine-tuning phase, we innovatively modified the loss function to enhance the model's precision in predicting coordinate values. These approaches ensure that our model possesses advanced cognitive and decision-making capabilities in driving scenarios. Consequently, our model achieved a score of 0.6064, securing the first prize on the competition's final results.
\end{abstract}

{\let\thefootnote\relax\footnote{{* Equal contributions}}}
{\let\thefootnote\relax\footnote{{$\dagger$ Primary contact: chenyuanpengcyp@gmail.com}}}

\section{Introduction}
Since the widespread adoption of ChatGPT, multimodal large language models have gained significant attention across various research fields due to their capability to process text and reason about non-textual data, such as images and videos. A substantial amount of research has also focused on applying large language models to the domain of autonomous driving. In this context, the competition aims to leverage the video comprehension abilities of large language models for cognitive and decision-making tasks in driving scenarios. By utilizing given multi-perspective images as input, the focus is on addressing questions related to perception, prediction, and planning within the autonomous driving system, ultimately assisting the vehicle in making interpretable driving decisions.

This technical report will detail our experimental approach, including the format of the input data, enhancements made during the pretraining and finetuning phases, and the final experimental results.

\section{Method}

\subsection{Multi View Image Input}
\begin{figure}
\centering
\includegraphics[width=1\linewidth]{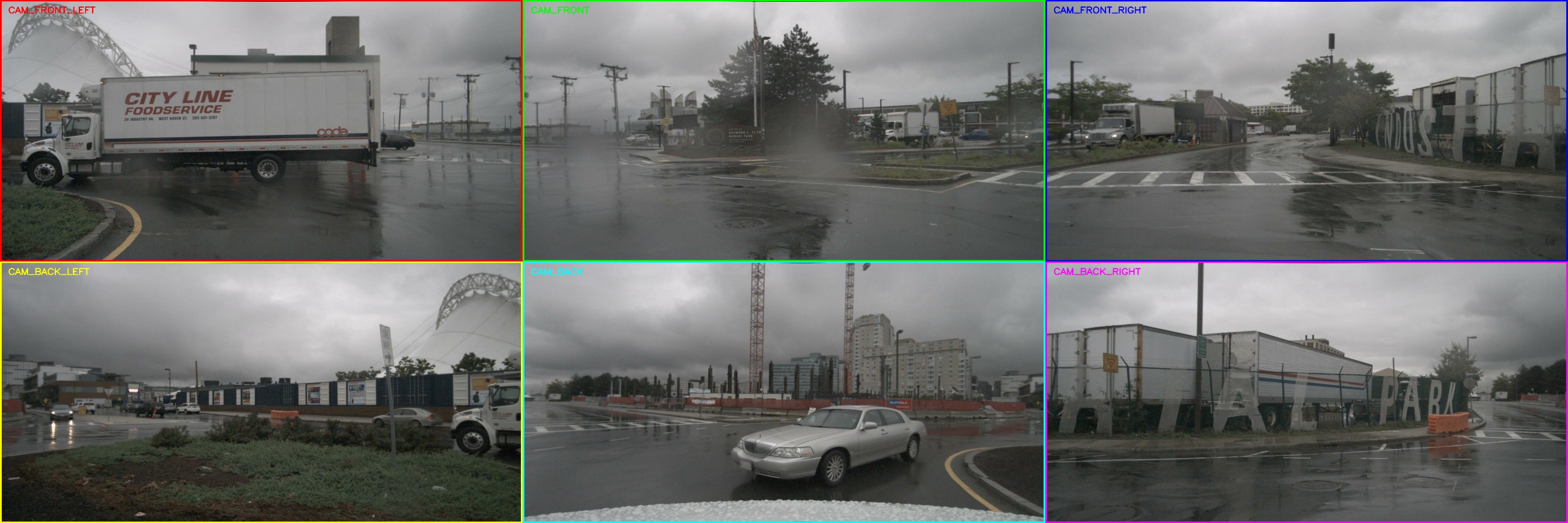}
\caption{\label{fig:input} Multi View Image.}
\end{figure}

Traditionally, Vision-Language Models (VLMs) process single-frame images sequentially, increasing training and inference costs and lacking inter-view contextual information. To adress these issues, we concatenated multi-view images into a single format, with each view resized to 896x448 pixels, resulting in a final image size of 2688x896 pixels, as shown in Fig \ref{fig:input}. We experimented with adjusting the target coordinates (\texttt{<C1, CAM FRONT, X, Y>}) in three ways: maintaining the original, resizing to match the individual view dimensions (896x448), and resizing to fit the concatenated image (2688x896). Our findings indicated that evaluation metrics are significantly affected by these adjustments, particularly the sensitivity of the language metric and match metric to the decimal and integer part of coordinate values. 
To prevent precision loss and optimize results, we opted to use the original target coordinates without any transformation.

\subsection{Training}
\begin{figure}
\centering
\includegraphics[width=0.8\linewidth]{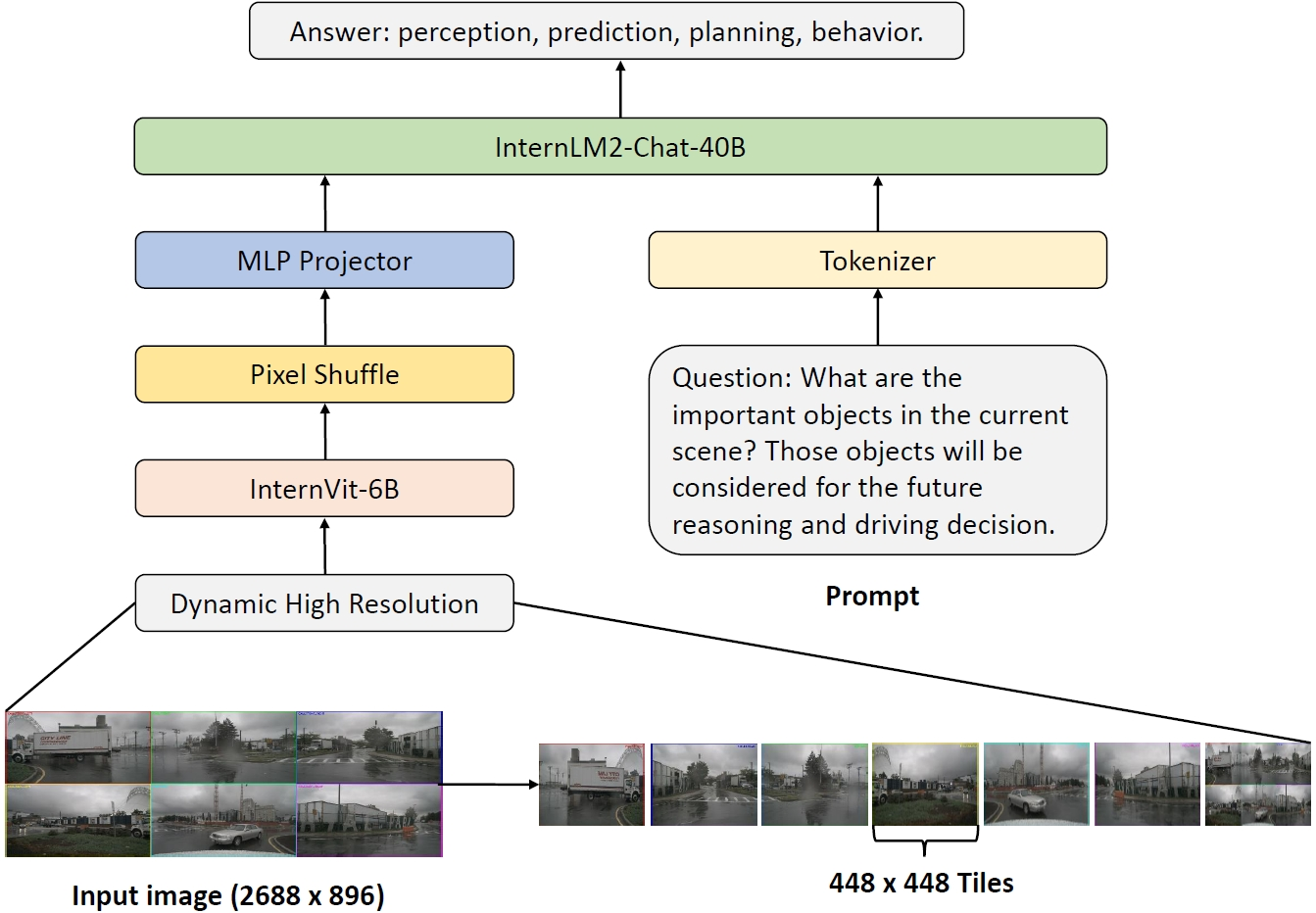}
\caption{\label{fig:model} Overall Architechture}
\end{figure}

We employed InternVL-2.0\cite{chen2024internvl}, a pioneering open-source multi-modal model as our base model. As shown in Fig \ref{fig:model}, the model splits the concatenated image into multiple high-resolution sub-images, which are used as input.

\textbf{System Prompt:} \textit{You are an Autonomous Driving AI assistant. You receive an image that consists of six surrounding camera views. The layout is as follows: The first row contains three images: FRONT LEFT, FRONT, FRONT RIGHT. The second row contains three images: BACK LEFT, BACK, BACK RIGHT. Your task is to analyze these images and provide insights or actions based on the visual data.}

\subsubsection{Pretraining}
To enhance the model's ability to detect and recognize important targets, we performed data augmentation while training since we discovered that the position of key objects \texttt{<C1, CAM FRONT, X, Y>}greatly influences the evaluation metrics. Specifically, we pre-trained the model on public autonomous driving datasets, including Nuscenes\cite{caesar2020nuscenes}, OpenLane-V2\cite{wang2024openlane}, Nuscenes-QA\cite{qian2024nuscenes}, Nuscenes-MQA\cite{inoue2024nuscenes}, and OmniDrive\cite{wang2024omnidrive}. All the targets relevant QA pairs from these datasets is sampled whereas they could be key or non-key objects. Moreover, we extracted 2D bounding boxes and 2D center coordinates of the targets in the Nuscenes and OpenLane-V2 datasets to enhance the model's grounding capabilities. For Nuscenes-QA, Nuscenes-MQA, and OmniDrive datasets with large volume of data, we compressed all QA pairs from the same frame into one QA pair to accelerate training. The overall usage of the dataset is presented in Table~\ref{tabel_1}.

\begin{table*}[h]\centering
\caption{Details of the Pre-trained Datasets.}
\vspace{0.2cm}
\renewcommand\arraystretch{2}
\begin{tabular}{l|l|l}
\hline
\textbf{Datasets}  & \textbf{Sampling Task Type} & \textbf{Number Of QA Pairs}\\ 
\hline
{Nuscenes}  & Grounding of obstacle     & 311430\\
\hline
{OpenLane-V2} & Grounding of traffic elements and load marks     & 40104\\
\hline
{Nuscenes-QA} & Perception     & 105642\\
\hline
{Nuscenes-MQ} & Perception     & 305740\\
\hline
{OmniDrive} & Perception, Prediction, Planning, Behavior   & 196403\\
\hline
{DriveLM} & Perception, Prediction, Planning, Behavior   & 377956\\
\hline
\end{tabular}
\label{tabel_1}
\end{table*}
\noindent

\subsubsection{Finetuning}
We incorporated positional constraints of the targets in the DriveLM dataset \cite{sima2023drivelm} to further enhance the model's fine-grained understanding of target positions. Specifically, we employed a loss function $Loss_{text}$ to calculate the language loss, and extracted the (X,Y) coordinates from the ground truth to compute $Loss_{Location}$, the overall loss is as follows (1).
\begin{equation}
    \mathit{Loss}_{\mathit{total}} = \lambda_1 \cdot \mathit{Loss}_{\mathit{text}} + \lambda_2 \cdot \mathit{Loss}_{\mathit{location}}
\end{equation}
Here, $Loss_{text}$ refers to the CrossEntropyLoss computed between the true label tokens and the predicted probability tokens. $Loss_{location}$ refers to the CrossEntropyLoss computed between the true label and the predicted tokens, which represent the coordinates of key targets. $\lambda_1$ and $\lambda_2$ are the weight coefficients for the loss. During the fine-tuning process, we exclusively used the DriveLM dataset.

\section{Experiments}
We utilized the InternVL 2.0 40B model and trained it with LoRA\cite{hu2021lora} for one epoch to achieve the final results. As shown in Table~\ref{tabel_2}, our method achieved a final score of 0.6064, exceeding DriveLM by 0.1079.

\begin{table*}[h]\centering
\caption{Performance Comparison of our method and DriveLM method.}
\vspace{0.2cm}
\renewcommand\arraystretch{2}
\resizebox{\textwidth}{!}{%
\begin{tabular}{l|*{10}{c}} 
\hline
\textbf{Method}  & \textbf{Accuracy}$\uparrow$ & \textbf{ChatGPT}$\uparrow$ & \textbf{Bleu\_1}$\uparrow$ & \textbf{Bleu\_2}$\uparrow$ & \textbf{Bleu\_3}$\uparrow$ & \textbf{Bleu\_4}$\uparrow$ & \textbf{ROUGE\_L}$\uparrow$ & \textbf{CIDEr}$\uparrow$ & \textbf{Match}$\uparrow$ & \textbf{Final Score}$\uparrow$ \\ 
\hline
{DriveLM}  & 0.6468 & 53.4192 & 0.7160 & 0.6526 & 0.5919 & 0.5324 & 0.7152 & 0.0788 & 32.8851 & 0.4985\\
\hline
{Ours} & \textbf{0.7625} & \textbf{64.1103} & \textbf{0.7633} & \textbf{0.7011} & \textbf{0.6425} & \textbf{0.5860} & \textbf{0.7405} & \textbf{0.2081} & \textbf{50.895} & \textbf{0.6064}\\
\hline
\end{tabular}
}
\label{tabel_2}
\end{table*}
\noindent

\subsection{Ablation Study}
To validate the effect of coordinates (\texttt{<C1, CAM FRONT, X, Y>}) in our model, we conducted ablation studies on the DriveLM dataset, as detailed in Table~\ref{tabel_3}, the model are trained using DriveLM dataset, InternVL 2.0 8B model by LoRA.
The results indicate that the match score for the original coordinates is 1.1\% higher than that for the concatenated coordinates. We hypothesize that this discrepancy arises from potential precision loss during the transformation between the original and new coordinates.
\begin{table*}[h]
\centering
\caption{\centering{The impact of coordinate types on the positions of key targets.}}
\vspace{0.2cm}
\renewcommand\arraystretch{2}
\resizebox{\textwidth}{!}{%
\begin{tabular}{l|*{10}{c}} 
\hline
\textbf{Coordinate Type}  & \textbf{Accuracy}$\uparrow$ & \textbf{ChatGPT}$\uparrow$ & \textbf{Bleu\_1}$\uparrow$ & \textbf{Bleu\_2}$\uparrow$ & \textbf{Bleu\_3}$\uparrow$ & \textbf{Bleu\_4}$\uparrow$ & \textbf{ROUGE\_L}$\uparrow$ & \textbf{CIDEr}$\uparrow$ & \textbf{Match}$\uparrow$ & \textbf{Final Score}$\uparrow$ \\ 
\hline
{Concatenated coordinates (2688x896 pixels)} & 0.7165 & \textbf{62.5512} & \textbf{0.7587} & \textbf{0.6976} & \textbf{0.6408} & \textbf{0.5856} & \textbf{0.7387} &0.1751 &42.0488 & \textbf{0.5727}\\
\hline
{Original coordinates} & \textbf{0.7419} &58.8181 &0.7370 &0.6763 &0.6197 &0.5651 &0.7312 & \textbf{0.1784} & \textbf{43.1858} &0.5632\\
\hline
\end{tabular}
}
\label{tabel_3}
\end{table*}
\noindent

\hspace*{-\parindent}In addition, we investigated the impacts of pretraining, finetuning, and $\text{Loss}_{{location}}$ strategies, along with the effects of model parameter size, as shown in Table~\ref{tabel_4}.

\begin{table*}[h]
\centering
\caption{\centering{Performance Comparison of Training Strategies, Model Parameters, and $\text{Loss}_{location}$}}
\vspace{0.2cm}
\renewcommand\arraystretch{2}
\resizebox{\textwidth}{!}{%
\begin{tabular}{c|c|c|c|*{10}{c}}
\hline
\textbf{Training Strategy} & \textbf{Model Parameters} & $\textbf{Loss}_{\text{location}}$  & \textbf{Datasets} & \textbf{Accuracy}$\uparrow$ & \textbf{ChatGPT}$\uparrow$ & \textbf{Bleu\_1}$\uparrow$ &\textbf{Bleu\_2}$\uparrow$ & \textbf{Bleu\_3}$\uparrow$ & \textbf{Bleu\_4}$\uparrow$ & \textbf{ROUGE\_L}$\uparrow$ & \textbf{CIDEr}$\uparrow$ & \textbf{Match}$\uparrow$ & \textbf{Final Score}$\uparrow$ \\ 
\hline
{InternVL + Finetune} & InternVL 2.0 8B & \textbf{$\boldsymbol{\times}$}  &DriveLM  & 0.7419  & 58.8181  & 0.7370 & 0.6763 & 0.6197 & 0.5651 & 0.7312 & 0.1784 & 43.1858 & 0.5632 \\
\hline
\multirow{6}{*}{Pretrain} & \multirow{6}{*}{InternVL 2.0 8B} & \multirow{6}{*}{\textbf{$\boldsymbol{\times}$}}  & DriveLM & \multirow{6}{*}{0.7349} & \multirow{6}{*}{63.4067} & \multirow{6}{*}{0.7434} & \multirow{6}{*}{0.6824} & \multirow{6}{*}{0.6255} &\multirow{6}{*}{0.5706} &\multirow{6}{*}{0.7399} & \multirow{6}{*}{0.1926} & \multirow{6}{*}{40.7884} & \multirow{6}{*}{0.5765} \\
& & & OpenLane-V2 & & & & & & & & & &\\
& & & Nuscenes-QA & & & & & & & & & &\\
& & & Nuscenes-MQ & & & & & & & & & &\\
& & & OmniDrive & & & & & & & & & &\\
& & & DriveLM & & & & & & & & & &\\
\hline
{Pretrain + Finetune} &InternVL 2.0 8B & \textbf{$\boldsymbol{\times}$}  &DriveLM  &0.7349 &63.7325 &0.7434 &0.6824 &0.6255 &0.5706 &0.7399 &0.1926 &41.2381 &0.5787\\
\hline
{Pretrain + Finetune} &InternVL 2.0 8B &\checkmark  &DriveLM  &0.7540 &63.9807 &0.7488 &0.6877 &0.6303 &0.5746 &0.7358 &0.1780 &43.1405 &0.5872 \\
\hline
{Pretrain + Finetune} &InternVL 2.0 40B &\checkmark &DriveLM &\textbf{0.7625} &\textbf{64.1103} &\textbf{0.7633} &\textbf{0.7011} &\textbf{0.6425} &\textbf{0.5860} &\textbf{0.7405} &\textbf{0.2081} &\textbf{50.895} &\textbf{0.6064} \\
\hline
\end{tabular}
}
\label{tabel_4}
\end{table*}

\bibliographystyle{alpha}
\bibliography{sample}

\newcommand{\etalchar}[1]{$^{#1}$}
\begin{thebibliography}{CWW{\etalchar{+}}24}

\bibitem[CBL{\etalchar{+}}20]{caesar2020nuscenes}
Holger Caesar, Varun Bankiti, Alex~H Lang, Sourabh Vora, Venice~Erin Liong, Qiang Xu, Anush Krishnan, Yu~Pan, Giancarlo Baldan, and Oscar Beijbom.
\newblock nuscenes: A multimodal dataset for autonomous driving.
\newblock In {\em Proceedings of the IEEE/CVF conference on computer vision and pattern recognition}, pages 11621--11631, 2020.

\bibitem[CWW{\etalchar{+}}24]{chen2024internvl}
Zhe Chen, Jiannan Wu, Wenhai Wang, Weijie Su, Guo Chen, Sen Xing, Muyan Zhong, Qinglong Zhang, Xizhou Zhu, Lewei Lu, et~al.
\newblock Internvl: Scaling up vision foundation models and aligning for generic visual-linguistic tasks.
\newblock In {\em Proceedings of the IEEE/CVF Conference on Computer Vision and Pattern Recognition}, pages 24185--24198, 2024.

\bibitem[HSW{\etalchar{+}}21]{hu2021lora}
Edward~J Hu, Yelong Shen, Phillip Wallis, Zeyuan Allen-Zhu, Yuanzhi Li, Shean Wang, Lu~Wang, and Weizhu Chen.
\newblock Lora: Low-rank adaptation of large language models.
\newblock {\em arXiv preprint arXiv:2106.09685}, 2021.

\bibitem[IYTY24]{inoue2024nuscenes}
Yuichi Inoue, Yuki Yada, Kotaro Tanahashi, and Yu~Yamaguchi.
\newblock Nuscenes-mqa: Integrated evaluation of captions and qa for autonomous driving datasets using markup annotations.
\newblock In {\em Proceedings of the IEEE/CVF Winter Conference on Applications of Computer Vision}, pages 930--938, 2024.

\bibitem[QCZ{\etalchar{+}}24]{qian2024nuscenes}
Tianwen Qian, Jingjing Chen, Linhai Zhuo, Yang Jiao, and Yu-Gang Jiang.
\newblock Nuscenes-qa: A multi-modal visual question answering benchmark for autonomous driving scenario.
\newblock In {\em Proceedings of the AAAI Conference on Artificial Intelligence}, volume 38 Issue 5, pages 4542--4550, 2024.

\bibitem[SRC{\etalchar{+}}23]{sima2023drivelm}
Chonghao Sima, Katrin Renz, Kashyap Chitta, Li~Chen, Hanxue Zhang, Chengen Xie, Ping Luo, Andreas Geiger, and Hongyang Li.
\newblock Drivelm: Driving with graph visual question answering.
\newblock {\em arXiv preprint arXiv:2312.14150}, 2023.

\bibitem[WLL{\etalchar{+}}24]{wang2024openlane}
Huijie Wang, Tianyu Li, Yang Li, Li~Chen, Chonghao Sima, Zhenbo Liu, Bangjun Wang, Peijin Jia, Yuting Wang, Shengyin Jiang, et~al.
\newblock Openlane-v2: A topology reasoning benchmark for unified 3d hd mapping.
\newblock {\em Advances in Neural Information Processing Systems}, 36, 2024.

\bibitem[WYJ{\etalchar{+}}24]{wang2024omnidrive}
Shihao Wang, Zhiding Yu, Xiaohui Jiang, Shiyi Lan, Min Shi, Nadine Chang, Jan Kautz, Ying Li, and Jose~M Alvarez.
\newblock Omnidrive: A holistic llm-agent framework for autonomous driving with 3d perception, reasoning and planning.
\newblock {\em arXiv preprint arXiv:2405.01533}, 2024.

\end{thebibliography}
\end{document}